\documentclass{article}

\usepackage{iclr2021_preprint,times}

\usepackage[utf8]{inputenc} % allow utf-8 input
\usepackage[T1]{fontenc}    % use 8-bit T1 fonts
\usepackage{hyperref}       % hyperlinks
\usepackage{url}            % simple URL typesetting
\usepackage{booktabs}       % professional-quality tables
\usepackage{amsfonts}       % blackboard math symbols
\usepackage{amsmath}       % blackboard math symbols
\usepackage{amsthm}
\usepackage{enumitem}
\usepackage{nicefrac}       % compact symbols for 1/2, etc.
\usepackage{microtype}      % microtypography
\usepackage{graphicx}
\usepackage{wrapfig}

\DeclareMathOperator*{\E}{\mathbb{E}}
\DeclareMathOperator{\mcoherence}{\mathnormal{m}-coherence}
\newtheorem{theorem}{Theorem}
\newtheorem{lemma}[theorem]{Lemma}
\newtheorem{corollary}{Corollary}[theorem]

\title{Making Coherence Out of Nothing At All:\\
Measuring the Evolution of Gradient Alignment}

\author{%
  Satrajit Chatterjee\\
  Google AI \\
  Mountain View, CA 94043 \\
  \texttt{schatter@google.com} \\
  \And
  Piotr Zielinski\\
  Google AI \\
  New York, NY 10011 \\
  \texttt{zielinski@google.com} \\
}

\iclrfinalcopy 

\begin{document}

\maketitle

\begin{abstract}

We propose a new metric ($\mcoherence$) to experimentally study the alignment of per-example gradients during training. Intuitively, given a sample of size $m$, $\mcoherence$ is the number of examples in the sample that benefit from a small step along the gradient of any one example on average. We show that compared to other commonly used metrics, $\mcoherence$ is more interpretable, cheaper to compute ($O(m)$ instead of $O(m^2)$) and mathematically cleaner. (We note that $\mcoherence$ is closely connected to gradient diversity, a quantity previously used in some theoretical bounds.)
Using $\mcoherence$, we study the evolution of alignment of per-example gradients in ResNet and Inception models on ImageNet and several variants with label noise, particularly from the perspective of the recently proposed Coherent Gradients (CG) theory that provides a simple, unified explanation for memorization and generalization [Chatterjee, ICLR 20]. Although we have several interesting takeaways, our most surprising result concerns memorization. Na\"ively, one might expect that when training with completely random labels, each example is fitted independently, and so $\mcoherence$ should be close to 1. However, this is not the case: $m$-coherence reaches much higher values during training (100s), indicating that over-parameterized neural networks find common patterns even in scenarios where generalization is not possible. A detailed analysis of this phenomenon provides both a deeper confirmation of CG, but at the same point puts into sharp relief what is missing from the theory in order to provide a complete explanation of generalization in neural networks.

\end{abstract}

% Inner Products and Cauchy Schwartz http://www.cs.tau.ac.il/~amir1/PS/TEMP/3-ipn.pdf 

\section{Introduction}

Generalization in neural networks trained with stochastic gradient descent (SGD) is not well-understood. For example, the generalization gap, i.e., the difference between training and test error depends critically on the dataset and we do not understand how. This is most clearly seen when we fix all aspects of training (e.g. architecture, optimizer, learning rate schedule, etc.) and vary only the dataset. In a typical experiment designed to test this, training on a real data set (e.g., ImageNet) leads to a relatively small generalization gap, whereas training on randomized data (e.g., ImageNet with random labels) leads to a much larger gap~\citep{Zhang17, Arpit17}. 

The mystery is that in both cases (real labels and random) the training accuracy is close to 100\% which implies that the network and the learning algorithm have sufficient effective capacity~\citep{Arpit17} to memorize the training sets, i.e., to fit an arbitrary mapping from the input images to labels. But, what then, is the mechanism that from among all the maps consistent with the training set, allows SGD to find one that generalizes well (when such a well-generalizing map exists)? 

This question has motivated a lot of work (see e.g. ~\citet{Zhang17,Arpit17,Bartlett17,Kawaguchi17,Neyshabur18,Arora18,Belkin19,Rahaman19}) but no satisfactory answer has emerged. As \citet{Nagarajan19} point out, traditional approaches based on uniform convergence may not suffice, and new ideas are needed. A promising line of attack is via algorithmic stability~\cite{Bousquet02}, but traditional stability analysis of SGD (e.g., \citet{Hardt16, Kuzborskij18}) does not account for the dataset, and without that, one cannot hope to get more than a vacuous bound.

Recently, a new approach called Coherent Gradients (CG) has been proposed that takes into account the training dataset in reasoning about stability~\citep{Chatterjee20, Zielinski20}. By analogy to Random Forests which also show dataset  dependent generalization, CG posits that neural networks try to extract commonality from the dataset during the training process. 

The key insight is that, since the overall gradient for a single step of SGD is the sum of the per-example gradients, it is strongest in directions that reduce the loss on multiple examples if such directions exist. Intuitively, at one extreme, if all the per-example gradients are aligned we get perfect  stability (since dropping an example doesn’t affect the overall gradient) and thus perfect generalization. At the other extreme, if all the per-example gradients are pairwise orthogonal, we get no stability (since dropping an example eliminates any descent down its gradient), and thus pure memorization. 

Thus CG provides a simple, unified explanation for both memorization and generalization. However, at the same time, CG leads to some basic empirical questions:

\begin{enumerate}

\item  {\em What does the alignment of per-example gradients, i.e., {\em coherence} look like in practice?}

As was noted in~\citet{Chatterjee20}, we expect a real dataset to have more coherence than a dataset with random labels, but how big is this difference quantitatively? Is coherence in the random label case like that in the pairwise orthogonal case described above? How does it vary with layer or architecture? 

\item {\em Is the coherence constant throughout training, or does it vary? If so, how?}

The key insight of CG (as described above) is a point-in-time observation, but in order to get a full picture of generalization we need to analyse the entire training trajectory. For example, one might imagine that as more and more training examples are fitted, coherence decreases, but is it possible for it to increase in the course of training?

\end{enumerate}

In this paper, we propose a new metric called $\mcoherence$ to experimentally study gradient coherence. The metric admits a very natural intuitive interpretation that allows us to gain insight into the questions above. 
While we confirm our intuitions in many cases, we also find some surprises.
These observations help us formulate more precisely what is missing from the CG explanation for generalization, and thus point the way to future work in this direction.

%\section{A New Metric and Relation to Previous Work}
\section{Prior Work on Metrics for Experimentally Measuring Coherence}
\label{sec:prev}

{\bf Pairwise Dot Product.} An obvious starting point to quantify the alignment or coherence of a set of gradients is their average pairwise dot product. Since this has a nice connection to the loss function, we start by reviewing the connection, and also set up notation in the process.

Formally, let $\mathcal{D}(z)$ denote the distribution\footnote{We would like to quantify gradient coherence for both populations and samples. Therefore, $\mathcal{D}$ can either be a population distribution (typically unknown) or a sample  (i.e., empirical) distribution.} of examples from a finite\footnote{We assume finiteness for simplicity  since it does not affect generality for practical applications.} set $Z$, and assume without loss of generality that ${\rm support}(\mathcal{D}) = Z$. 
For a network with $d$ trainable parameters, 
let $\ell_z(w)$ be the loss for an example $z \sim \mathcal{D}$ for a parameter vector $w \in \mathbb{R}^d$. For the learning problem, we are interested in minimizing the expected loss $\ell(w) := \E_{z \sim \mathcal{D}}[\ell_z(w)]$.
Let $g_z := [\nabla \ell_z](w)$ denote the gradient of the loss on example $z$, and $g := [\nabla \ell](w)$ denote the overall gradient. From linearity, we have,
\[
g = \E_{z \sim \mathcal{D}}\ [\ g_z\ ]
\]
Now, suppose we take a small descent step $h = - \eta g$ (where $\eta > 0$ is the learning rate). From the Taylor expansion of $\ell$ around $w$, we have, 
\begin{equation}
\ell(w + h) - \ell(w) \approx g \cdot h 
= - \eta\  g \cdot g
= - \eta \E_{z \sim \mathcal{D}}\ [\ g_z\ ] \cdot \E_{z \sim \mathcal{D}}\ [\ g_z\ ]
= - \eta \E_{z \sim \mathcal{D}, z' \sim \mathcal{D}}\ [g_z \cdot g_{z'}]
%= - \eta \E_{z \sim \mathcal{D}}\ [\ g_z\ \cdot  g\ ]
\label{eq:dloss}
\end{equation}
where the last equality can be checked with a direct computation.
%
%TODO: potentially kill
Thus, the following are approximately equivalent:
\begin{itemize}
    \item reduction in loss (due to a small step) divided by the learning rate,
    \item squared $\ell^2$ norm of the expected gradient, and,
    \item expected pairwise dot product (where the expectation is over {\em all} pairs).
\end{itemize}

{\bf Example.}~\citep{Chatterjee20} Consider a sample with $m$ examples $z_i$ where $1 \leq i \leq m$. Let $g_i$ be the gradient of $z_i$ and further that
$\|g_i\| = \|u\|$ for some $u$.
If all the $g_i$ are the same, then $g \cdot g = \|u\|^2$. However, if they are pairwise orthogonal, i.e., $g_i \cdot g_j = 0$ for $i \neq j$, then $g \cdot g = \frac{1}{m} \|u\|^2$.
\qed

%\vspace{5mm}
As this illustrates, the average expected dot product can vary significantly depending on the coherence. However, as a metric for coherence it is rather fragile. For example, just re-scaling the loss can drastically alter the value of the metric. 
Therefore, it can only be used to reason about coherence in very limited settings. 
For e.g., \citet{Chatterjee20, Zielinski20} use it to verify that adding increasing amounts of label noise to a dataset reduces coherence but in order to do so they keep everything else the same, and limit their considerations to the start of training. 
But, to study the evolution of coherence, even over a single training run requires normalization since the magnitude of the gradients changes significantly in the course of training (e.g., see Appendix). 

{\bf Stiffness.} \citeauthor{Fort19} in their preprint \citeyearpar{Fort19} study two variants of the average pairwise dot product that they call {\em sign stiffness} and {\em cosine stiffness}. In our notation these are
\small
\[
S_{\rm sign} := 
\E_{\substack{z \sim \mathcal{D}, z' \sim \mathcal{D}\\ z \neq z'}}[\ {\rm sign}(g_z \cdot g_{z'})\ ]
\ \ 
{\rm and}
\ \ 
S_{\rm cos} := 
\E_{\substack{z \sim \mathcal{D}, z' \sim \mathcal{D}\\ z \neq z'}}\left[ \ \frac{g_z}{\|g_z\|} \cdot \frac{g_{z'}}{\|g_{z'}\|}\ \right].
\]
\normalsize
These are meant to capture how a small gradient step based on one input example affects the loss on a {\em different} input example. 
Although \citeauthor{Fort19} do not describe why they choose to transform the gradients in these specific ways, we expect it is to normalize the dot product so that it can be tracked in the course of training. 
In their experience, they found sign stiffness to be more useful to analyze stiffness between classes whereas cosine stiffness was more useful within a class.

{\bf Gradient Confusion.} \citeauthor{Sankararaman19} in their preprint \citeyearpar{Sankararaman19} introduce the notion of a gradient confusion bound. The {\em gradient confusion bound} is $\zeta \geq 0$ if for all
$z, z' \in Z$ and $z \neq z'$, we have,
$g_z \cdot g_{z'} \geq -\zeta$.
They use this concept to study theoretically the convergence rate of gradient descent, but in their experimental results they measure the minimum cosine similarity between gradients, i.e., 
\small
\[
\min_{\substack{z \in Z, z' \in Z \\ z \neq z'}} 
\left[ \ \frac{g_z}{\|g_z\|} \cdot \frac{g_{z'}}{\|g_{z'}\|}\ \right]
\]
\normalsize
We note that the non-linearities (and to a lesser extent the $z \neq z'$ restriction) make it hard to tie stiffness or minimum cosine similarity to what happens during training; specifically, to the change in the loss function as a result of a gradient step which is the expectation over {\em all} per-example gradients.

\section{A New Metric for Coherence}
\label{sec:new}

The key insight behind our proposal is that there is a natural scaling factor that can be used to normalize the expected dot product of per-example gradients (i.e., the quantity in (\ref{eq:dloss})) that preserves the connection to the loss.
Consider the Taylor expansion of each individual loss $\ell_z$ around $w$ when we take a small step $h_z$ down {\em its} gradient $g_z$:
\[
\ell_z(w + h_z) - \ell_z(w) \approx g_z \cdot h_z 
= - \eta\  g_z \cdot g_z
\]
Taking expectations over $z$ we get,
\begin{equation}
\E_{z \sim \mathcal{D}} [ \ell_z(w + h_z) - \ell_z(w) ] 
= - \eta\ \E_{z \sim \mathcal{D}} [ g_z \cdot g_z ]
\label{eq:dloss_ez}
\end{equation}
The quantity in (\ref{eq:dloss_ez}) has a simple interpretation: It is the reduction in the {\em overall} loss $\ell$ if each example $\ell_z$ could be optimized independently. 
As might be expected intuitively, it is an upper bound on the quantity in  (\ref{eq:dloss}) and is tight when all the per-example gradients are identical. We prove this formally in \S\ref{sec:facts}.
Thus, it serves as a natural scaling factor for the expected dot product, and we obtain a normalized metric for coherence (denoted by $\alpha$) from (\ref{eq:dloss}) and (\ref{eq:dloss_ez}): 
% 
%\scalebox{\textwidth}{
\small
\begin{equation}
\boxed{
%\quad
\ 
\alpha := 
\frac{\ell(w + h) - \ell(w)}{\displaystyle \E_{z \sim \mathcal{D}}\ [ \ell_z(w + h_z) - \ell_z(w) ]} =
\frac{\displaystyle \E_{z \sim \mathcal{D}, z' \sim \mathcal{D}}\ [g_z \cdot g_{z'}]}
{\displaystyle \E_{z \sim \mathcal{D}}\ [ g_z \cdot g_z ]}
=
\frac{\displaystyle \E_{z \sim \mathcal{D}}\ [\ g_z\ ] \cdot \E_{z \sim \mathcal{D}}\ [\ g_z\ ]}
{\displaystyle \E_{z \sim \mathcal{D}}\ [ g_z \cdot g_z ]}
=
\frac
{\displaystyle \E_{z \sim \mathcal{D}}\ [\ g_z \cdot g_{\ } ]}
{\displaystyle \E_{z \sim \mathcal{D}}\ [\ g_z \cdot g_z ]}
%\quad
\ 
}
\label{eq:metric}
\end{equation}
\normalsize
%}
% 
% where the last equality follows from a direct computation:
% \[
% \E_{z \sim \mathcal{D}, z' \sim \mathcal{D}}\ [g_z \cdot g_{z'}]
% =
% \E_{z \sim \mathcal{D}}\ [\ g_z\ ] \cdot \E_{z \sim \mathcal{D}}\ [\ g_z\ ]
% =
% \E_{z \sim \mathcal{D}}\ \left[\ g_z\ \cdot \E_{z \sim \mathcal{D}}\ [\ g_z\ ] \right].
% \]

{\em Thus, $\alpha$ is the change in the overall loss due to a small gradient step as a fraction of the maximum possible change in loss if each component of the loss could be optimized independently}. 

As noted before, $0 \leq \alpha \leq 1$, and the maximum is achieved when all the gradients are identical, and the minimum is achieved when the expected gradient is 0, i.e., a stationary point is reached. 

{\bf A natural scale for $\alpha$.} 
Once again, consider a sample with $m$ examples $z_i$ where $1 \leq i \leq m$. Let $g_i$ be the gradient of $z_i$. Suppose further that the $g_i$ are pairwise orthogonal i.e. $g_i \cdot g_j = 0$ for $i \neq j$.
It is easy to check that $\alpha = 1/m$. For a sample of size $m$, we call this value of $\alpha$ the {\em orthogonal limit}. 

Since in the orthogonal case, each example is optimized independently, going down the expected gradient is $1/m$ times as slow as optimizing each independently. If the gradients are better aligned, we expect them to help each other resulting in an $\alpha$ greater than the orthogonal limit.

{\bf Example (Commonality).} For $1 \le i \le m$, suppose each $g_i$ has a common component $c$ and an idiosyncratic component $u_i$, i.e., $g_i = c + u_i$ with $u_i \cdot u_j = 0$ for $1 \le j \le m$ and $j \ne i$; $u_i \cdot c = 0$; and say, $u_i \cdot u_i = \|u\|^2$ for some $u$. It is easy to see that $\alpha$ in this case is 
$ \frac{1}{m} \left[ 1 + (m - 1) \cdot f  \right] $
where $f = \|c\|^2 / (\|c\|^2 + \|u\|^2)$.
\qed

These examples along with the observation that $0 \le \alpha \le 1$ suggests a more evocative (even if less accurate and less general) interpretation: In a given sample, $\alpha$ is the average fraction of examples that each example helps or supports.
Thus, when analyzing experimental data, for a sample of size $m$, it is convenient to define a new quantity $\mcoherence$ as follows:
\small
\[
\boxed{
\qquad
\mcoherence := m \cdot \alpha 
=
m \cdot \frac{\displaystyle \E_{z \sim \mathcal{D}, z' \sim \mathcal{D}}\ [g_z \cdot g_{z'}]}
{\displaystyle \E_{z \sim \mathcal{D}}\ [ g_z \cdot g_z ]}
\qquad
}
\]
\normalsize
Thus $m$-coherence in the orthogonal limit is 1 and in the identical case is $m$. {\em Intuitively, $m$-coherence of a sample is the number of examples (including itself) that any one example helps on average.}

{\bf Advantages.} $\alpha$ and $\mcoherence$ have several advantages over the metrics discussed in \S\ref{sec:prev}:

\begin{itemize}%[left=1em]

\item {\bf Computational Efficiency.} For a sample of size $m$, due to (\ref{eq:metric}), $\alpha$ can be computed exactly in $O(m)$ time in contrast to $O(m^2)$ time required for stiffness and cosine dot products. Furthermore, it can be computed in a streaming fashion by keeping two running sums, so the per-example gradients need not be stored. Thus, in our experiments we are able to use sample sizes a couple of orders of magnitude higher than those in~\citet{Fort19} and \citet{Sankararaman19}. 

\item {\bf Mathematical Simplicity.} We believe our definition is cleaner mathematically. This allows us to reason about the metric more easily. For example,

\begin{enumerate}

    \item We can show that the coherence of minibatch gradients is greater than that of individual examples (Corollary~\ref{cor:amp}). Therefore, care must be taken
    if minibatch gradients are used in lieu of example gradients in computing coherence (e.g. as in \citet{Sankararaman19}).

    \item Explicitly ruling out $z \neq z'$ as in done in stiffness and cosine similarity to eliminate  self-correlation is unnatural and can get tricky in practice due to near-duplicates or multiple examples leading to same or very similar gradients. 
    We obtain meaningful values without imposing those conditions, but if one insists on removing self-correlations, then subtracting $1/m$ from $\alpha$ or 1 from $m$-coherence is a more principled way to do it. 
    
    \item The non-linearities in stiffness and cosine similarity amplify small per-example gradients potentially overstating their importance, and lead to a discontinuity (or undefined behavior) with zero gradients. However, we can cleanly account for the effect of 
    negligible gradients in our observations (e.g. see Lemma~\ref{lem:zero}).
    
\end{enumerate}

\item {\bf Interpretability.} Finally, as discussed in detail above, they are normalized and yet easily interpretable due to the natural connection with loss.
    
\end{itemize}

{\bf Prior Work on Gradient Diversity.} While writing this paper we discovered that the reciprocal of $\alpha$ appears in the theory literature as {\em gradient diversity}. This was used by~\cite{Yin18} in theoretical bounds to understand the effect of mini-batching on convergence of SGD. (A similar result appears for least squares regression in \citet{Jain18}.) 
They show that the greater is the gradient diversity, the more effective are large mini-batches in speeding up SGD. Although they support their theoretical analysis with experiments on {\sc cifar}-10 (where they replicate $1/r$ of the dataset $r$ times and show that greater the value of $r$ less the effectiveness of mini-batching to speed up) they never actually measure the gradient diversity in their experiments (or further study its properties).
Also, note that for our purposes $\alpha$ is a better choice than $1/\alpha$ -- not just because coherence rather than incoherence is 
what leads to generalization -- but also since the latter can diverge: $g$ can be 0 without all $g_z$ being zero (e.g. at the end of training in an under-parameterized setting).

% plot n * x / (1 + (n - 1)x), x =0 to 1, n = 4096

\section{A More General Setting for Coherence and Some Basic Facts}
\label{sec:facts}

Our notion of coherence is not specific to gradients (or optimization) but extends naturally to vectors in Euclidean spaces. 
%
% Here, we list some basic facts about it in this more general setting (the proofs are in the supplement).
% 
Let $\mathcal{V}$ be a probability distribution on a collection of $m$ vectors in an Euclidean space.
In accordance with (\ref{eq:metric}), we define the {\em coherence} of $\mathcal{V}$ (denoted by  $\alpha(\mathcal{V})$) to be
\small
\begin{equation}
\alpha(\mathcal{V}) = \frac{\displaystyle \E_{v \sim \mathcal{V}, v' \sim \mathcal{V}} \  [ v \cdot v' ]}{\displaystyle \E_{v \sim \mathcal{V}} \  [ v \cdot v]}
%\alpha(V) = \frac{\E_{v, v' \sim \mathcal{V}} [ v \cdot v' ]}{\E_{v \sim \mathcal{V}} [ v \cdot v]}
%\alpha_{\mathcal{V}}(\mathbb{V}) = \frac{\E [ V \cdot V' ]}{\E [ V \cdot V]}
\label{eq:definealpha}
\end{equation}
\normalsize
Note that $\E[v \cdot v] = 0$ implies $\E[v \cdot v'] = 0$.
In what follows, we ignore the technicality of the denominator being 0 by always assuming that there is at least one non-zero vector in the support of $\mathcal{V}$ (which also held in our experiments). 
We list some basic facts. % The proofs are in the supplement. 

\vspace{5mm}
\begin{theorem}[Boundedness]
We have $0 \leq \alpha(\mathcal{V}) \leq 1$. In particular, $\alpha(\mathcal{V}) = 0$ iff $\E_{v \sim \mathcal{V}} [v] = 0$ and $\alpha(\mathcal{V}) = 1$ iff all the vectors are equal.
\end{theorem}

\begin{proof}
Since $v \cdot v \geq 0$ for any $v$, we have $\E_{v \sim \mathcal{V}}[v \cdot v] \geq 0$. Furthermore, it is easy to verify by expanding the expectations (in terms of the vectors and their corresponding probabilities) that
\begin{equation}
\displaystyle \E_{v \sim \mathcal{V}, v' \sim \mathcal{V}} \  [ v \cdot v' ] = \E_{v \sim \mathcal{V}} [v]\ \cdot \E_{v \sim \mathcal{V}} [v] 
%= \left\Vert \E_{v \sim \mathcal{V}} [v] \right\Vert^2 
\geq 0.
\label{eq:norm}
\end{equation}
Therefore, $\alpha(\mathcal{V}) \geq 0$.
Likewise, another direct computation shows that
\begin{equation}
%\displaystyle 
%
0 \leq 
\E_{v' \sim \mathcal{V}}\ \left[
  (\E_{v \sim \mathcal{V}}[v] - v')
        \cdot 
  (\E_{v \sim \mathcal{V}}[v] - v') 
\right] 
= 
% \E_{v' \sim \mathcal{V}}\ \left[
%   \E_{v \sim \mathcal{V}}[v] \cdot \E_{v \sim \mathcal{V}}[v] 
%   - 2 \E_{v \sim \mathcal{V}}[v] \cdot v' 
%   + v' \cdot v'
% \right] \\
%& =
\E_{v \sim \mathcal{V}}[v \cdot v] -
\E_{v \sim \mathcal{V}}[v]\ \cdot \E_{v \sim \mathcal{V}}[v] 
\end{equation}
Since from Equation~\ref{eq:norm} we have $\E[v] \cdot \E[v] = \E[v \cdot v']$, it follows that $\alpha(\mathcal{V}) \leq 1$.
Furthermore, since each term of the expectation on the left is
non-negative, equality is attained only when all the vectors are equal.
\end{proof}

\vspace{5mm}
\begin{lemma}[Scale Invariance]
For non-zero $k \in \mathbb{R}$, 
let $k \mathcal{V}$ denote the distribution of the random variable $k v$ where $v$ is drawn from $\mathcal{V}$.
We have $\alpha(k \mathcal{V}) = \alpha(\mathcal{V})$.
\label{lem:scale}
\end{lemma}

\begin{proof}
\begin{equation}
\alpha(k\mathcal{V}) 
= \frac{\displaystyle \E_{v \sim k \mathcal{V}, v' \sim k \mathcal{V}} \  [ v \cdot v' ]}{\displaystyle \E_{v \sim k \mathcal{V}} \  [ v \cdot v]}
= \frac{\displaystyle \E_{v \sim \mathcal{V}, v' \sim \mathcal{V}} \  [ k v \cdot k v' ]}{\displaystyle \E_{v \sim \mathcal{V}} \  [ k v \cdot k v]}
= \frac{\displaystyle \E_{v \sim \mathcal{V}, v' \sim \mathcal{V}} \  [ v \cdot v' ]}{\displaystyle \E_{v \sim \mathcal{V}} \  [ v \cdot v]}
= \alpha(\mathcal{V})
\end{equation}
\end{proof}

% {\bf Orthogonal vectors.} Now suppose
% $v_i \cdot v_j = 0$ for $i \neq j$ ($1 \leq i, j \leq m$)
% and let $\mathcal{V}_{\perp}$ be the uniform distribution on
% $\{v_1, v_2, .., v_m\}$. In this particular case, it is easy to check from a direct computation that $\alpha(\mathcal{V}_{\perp}) = \frac{1}{m}$.

\vspace{5mm}
\begin{theorem}[Stylized mini-batching]
Let $v_1, v_2, .., v_k$ be $k$ i.i.d. variables drawn from $\mathcal{V}$. Let $\mathcal{W}$ denote the distribution of the random variable $w = \frac{1}{k} \sum_{i=1}^{k} v_i$. We have,
\small
\begin{equation}
    \alpha(\mathcal{W}) = 
    \alpha(k \mathcal{W}) = 
    \frac{k \cdot \alpha(\mathcal{V})}{1 + (k - 1) \cdot \alpha(\mathcal{V})}
\end{equation}
\normalsize
\end{theorem}

\begin{proof}
The first equality follows from Lemma~\ref{lem:scale}. For the second equality, we have,
\begin{equation*}
\alpha(k \mathcal{W}) = 
\frac{\displaystyle \E_{\substack{w \sim k \mathcal{W}, \\ w' \sim k \mathcal{W}}} \  [ w \cdot w' ]}{\displaystyle \E_{w \sim k \mathcal{W}} \  [ w \cdot w]} =
\frac{\displaystyle \E_{\substack{v_1, .., v_k, \\ v_1', .., v_k'}} \  [ (\sum_i v_i) \cdot (\sum_i v_i') ]}{\displaystyle \E_{v_1, .., v_k} \  [(\sum_i v_i) \cdot (\sum_i v_i)]} =
\frac{\displaystyle k^2 \E_{\substack{v \sim \mathcal{V},\\v' \sim \mathcal{V}}} \  [ v \cdot v' ]}
{\displaystyle k \E_{v \sim \mathcal{V}} \  [ v \cdot v] + k\cdot(k - 1)\E_{\substack{v \sim \mathcal{V},\\v' \sim \mathcal{V}}} \  [ v \cdot v' ]} 
\end{equation*}
By dividing the numerator and denominator of the last expression by $\displaystyle k \E_{v \sim \mathcal{V}} \ [v \cdot v]$ the required result follows.
\end{proof}

\vspace{5mm}
\begin{corollary}[Minibatch amplification]
\label{cor:amp}
$\alpha(\mathcal{W}) \geq \alpha(\mathcal{V})$ with equality iff $\alpha(\mathcal{V}) = 0$ or $\alpha(\mathcal{V}) = 1$.
\end{corollary}

\begin{proof}
From the previous theorem, the transformation in coherence due to stylized mini-batching is given by the map $\alpha \mapsto \frac{k \cdot \alpha}{1 + (k - 1) \cdot \alpha}$. 
Now, since $\alpha \leq 1$, we have $k \geq 1 + (k - 1) \cdot \alpha$, and since $\alpha \geq 0$, multiplying both sides by $\frac{\alpha}{1 + (k - 1) \cdot \alpha}$ we have $\frac{k \cdot \alpha}{1 + (k - 1) \cdot \alpha} \geq \alpha$. Finally, it is easy to check that the only two fixed points of the map are $\alpha = 0$ and $\alpha = 1$.
\end{proof}

{\bf Remark.} This formulation provides a nice perspective on the type of results proved in \citet{Yin18} and \citet{Jain18}. When
$\alpha \ll 1/k$ but non-zero (i.e., we have high gradient diversity), creating mini-batches of size $k$ increases coherence almost $k$ times. But, when  $\alpha \approx 1$ (i.e., low diversity) there is not much point in creating mini-batches since there is little room for improvement.

\vspace{5mm}
\begin{lemma}[Effect of zero gradients]
\label{lem:zero}
If $\mathcal{W}$ denotes the distribution where with probability $p > 0$ we pick a vector from $\mathcal{V}$ and with probability $1 - p$ we pick the zero vector then $\alpha(\mathcal{W}) =  p \cdot \alpha(\mathcal{V})$.
% \begin{equation}
%     \alpha(\mathcal{W}) =  p \cdot \alpha(\mathcal{V})
% \end{equation}
\end{lemma}

\begin{proof}
\begin{equation}
\alpha(\mathcal{W}) 
= \frac{\displaystyle \E_{w \sim \mathcal{W}, w' \sim \mathcal{W}} \  [ w \cdot w' ]}{\displaystyle \E_{w \sim \mathcal{W}} \  [ w \cdot w]}
= \frac{\displaystyle p^2 \cdot \E_{v \sim \mathcal{V}, v' \sim \mathcal{V}} \  [ v \cdot v' ]}{\displaystyle p \cdot \E_{v \sim \mathcal{V}} \  [ v \cdot v]}
%= \frac{\displaystyle \E_{v \sim \mathcal{V}, v' \sim \mathcal{V}} \  [ v \cdot v' ]}{\displaystyle \E_{v \sim \mathcal{V}} \  [ v \cdot v]}
= p \cdot \alpha(\mathcal{V})
\end{equation}
\end{proof}

{\bf Example (Coherence Reduction).} If we add $k$ zero gradients to the collection of gradients constructed in 
the example of \S\ref{sec:new} (Commonality), using Lemma~\ref{lem:zero}, we get,
\small
\[
\alpha = \frac{m}{m + k} \cdot \frac{1}{m} \left[ 1 + (m - 1) \cdot f  \right] 
= \frac{1}{n} \left[ 1 + (n - k - 1) \cdot f  \right]
\]
\normalsize
where $n = m + k$ is the size of this new sample. For a fixed $n$, as $k$ increases, $\alpha$ decreases going down to $1/n$ (the orthogonal limit) when all but one vector in the sample is zero,
%
%\footnote{As noted before the all zero case is undefined though this line of reasoning suggests 0 may be a convenient definition.}
%
i.e., $k = n - 1$.

\section{Experimental Results}

\begin{figure}[t]
\centering
\includegraphics[width=1\textwidth]{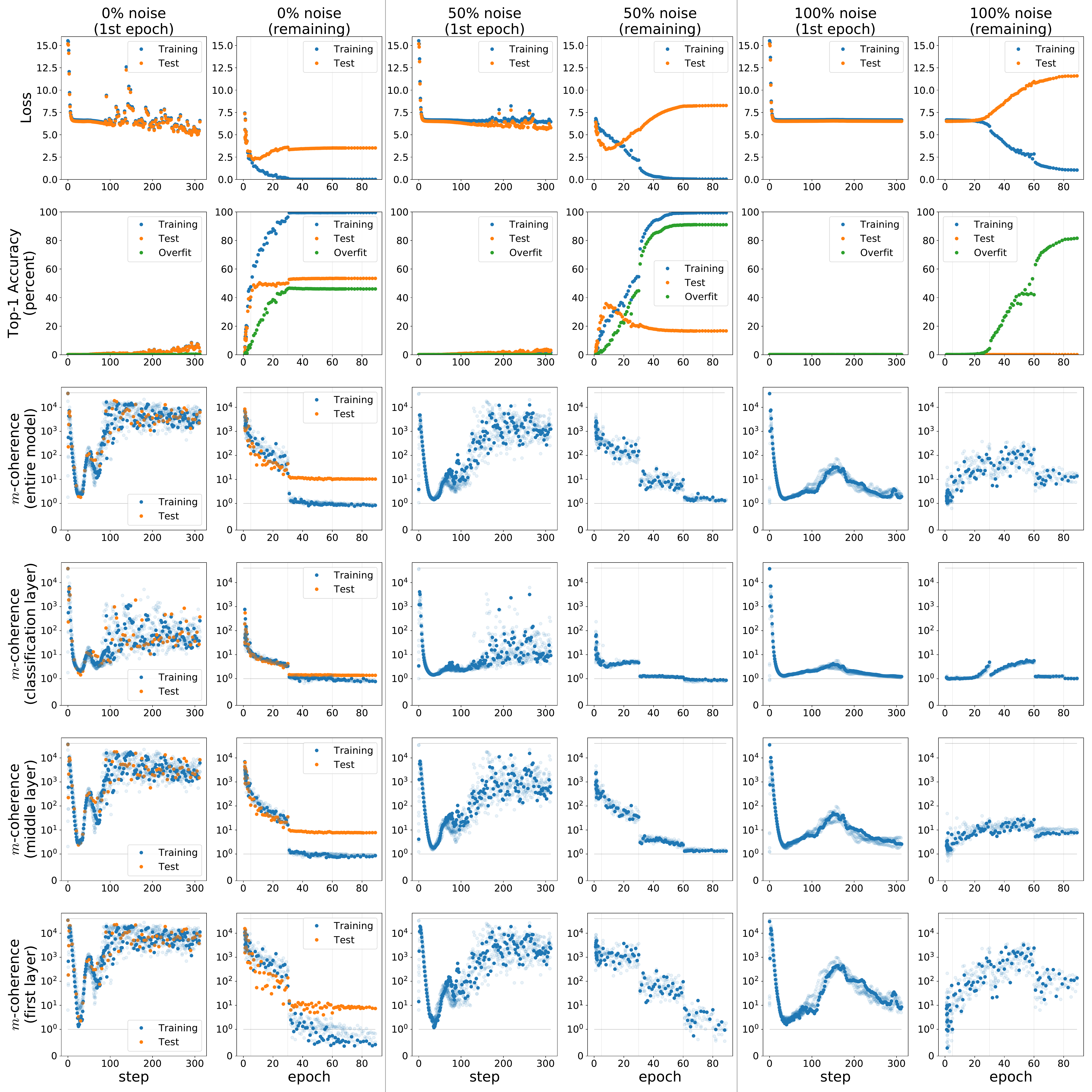}
\vskip -0.1in
\caption{The evolution of alignment of per-example gradients of a ResNet-18 during training as measured with $\mcoherence$ on samples of size $m=40,356$ on 3 variants of ImageNet with different amounts of label noise. Our main finding is that coherence not only decreases in the course of training (as might be expected when examples get fit), but it also increases. The peak is reached rapidly with real labels (within the first 100 steps) and slowly with random labels (over many epochs). Horizontal lines for $m$-coherence are shown at 1 (the orthogonal limit) and at $m$. Vertical lines indicate sharp reductions in learning rate. Light dots show the results of 4 other runs to understand sensitivity w.r.t. randomness in initialization and mini-batch construction.}
\label{fig:resnet_full}
\vspace*{-0.2in}
\end{figure}

We now use $\mcoherence$ to experimentally study the evolution of coherence.

{\bf Methodology.} We train ResNet-18 models on ImageNet with original labels (0\% noise), and
two derived datasets: one with half the training labels randomized (50\% noise), and another with all the training labels randomized (100\% noise).\footnote{We use the original ImageNet validation set as our test set in all cases.} We using SGD with momentum (0.9), a batch size of 4096, and the learning rate schedule proposed in~\cite{Goyal17}.
% gradual warm-up for the first 5 epochs, reduction of learning rate by the factor of 1/10th at 30th, 60th and 80th epoch.
%
We turn off augmentation and weight decay to observe memorization in the noisy cases within a reasonable number of steps.
For each dataset, we track $\mcoherence$ on a random (but fixed) set of $m = 40,356$ {\em training} examples.

Figure~\ref{fig:resnet_full} shows the data from our experiments. Each column corresponds to a different experiment and the rows show loss, accuracy, and $\mcoherence$ for the entire model and some specific layers.

{\bf Real Labels.} 
Our first experiment (shown in the {\em second} column of Figure~\ref{fig:resnet_full} for reasons that will become clear shortly) measures the $\mcoherence$ (row 3) for training with 0\% noise, i.e., the real ImageNet labels.
The initial coherence in epoch 1 is very high, almost $10^4$ and it decreases as more training examples get fit. 
We note that although there is some fluctuation in the coherence, it stays high (above $10^2$ and often above $10^3$) until well after the accuracy crosses the 50\% mark.
It settles at 1 after all the examples are fit.  

The high initial coherence agrees well with the intuition from CG that real datasets have good per-example gradient alignment since that is what is necessary for good generalization as per the theory. 
The subsequent decrease in coherence in the course of training is expected from Lemma~\ref{lem:zero} under the assumption that the gradients of fitted examples become small.

{\bf Random Labels.}
Our second experiment (column 6) shows that with random labels, the initial coherence in epoch 1 is low (between 1 and 10). It increases steadily until it reaches a peak in epochs 40 to 60 (between $10^2$ and $10^3$) followed by a decrease.

The low initial coherence (near the orthogonal limit) agrees well with CG as discussed in the introduction but the subsequent increase is surprising (though not in contradiction with CG as discussed later). The increase is not small since at its peak each example is helping hundreds of other examples (though it is well below the peak seen with real labels).
Once again, as examples get fitted, coherence decreases as expected from Lemma~\ref{lem:zero}, though not back down to 1, likely since our training only goes on till about 80\% accuracy is reached.

The increase in one case and not the other leads to a natural question with implications about the dynamics of SGD:
{\em Is the evolution of coherence fundamentally different between the well-generalizing case (real labels) and the memorization case (random labels)?}

{\bf Early Training.} To study this question, we took a closer look at the 1st epoch. We recorded $\mcoherence$ at initialization (i.e., before the first step) and, thereafter, for every step in the epoch. Since this requires computing the per-example gradients for $\approx$ 40K examples after every step, this was our most computationally expensive experiment taking 2-3 days per run (using TPUs).
The results are shown in columns 1 (real labels) and 5 (random).

% This is very finicky: even moving it up and down by a para screws things up. It interacts badly with section breaks and top/bottom of page.

\begin{wrapfigure}{R}{0.65\textwidth}
\centering
\includegraphics[width=0.5\textwidth]{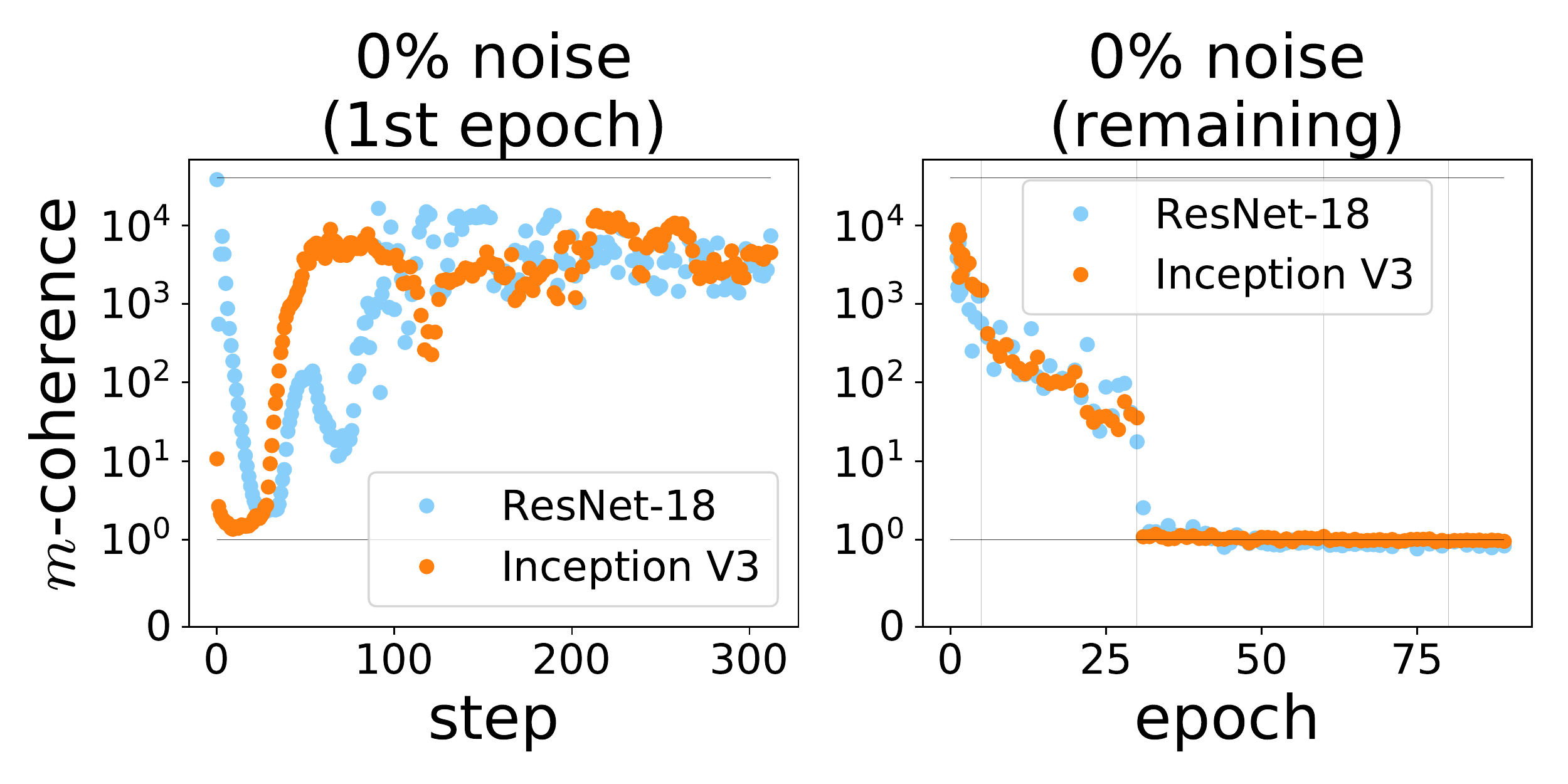}
%\vskip -0.5in
\caption{The early trajectory of an Inception model shows a similar (post-transient) increase as ResNet. The overall trajectories are also similar, and it is interesting to note that we
get similar values for coherence although the two architectures are very different.}
\label{fig:inception}
\vspace*{-0.1in}
\end{wrapfigure}

In the first 25 steps, in {\em both} cases, we find that coherence shows a steep fall from nearly $m$ down to less than 10. This is accompanied by a sharp decrease in training loss (cross-entropy) (row 1) from around 16 to ${\rm ln}(1000) \approx 6.9$ (the value expected from a uniform distribution at the outputs). 

However, after this initial transient, we find that in both cases the coherence starts from the low point and starts rising. The rise is much faster for real labels than for random labels. We ran additional experiments with 50\% noise (column 3) and with 25\% and 75\% noise (see Appendix) to confirm that the slope with which coherence increases depends inversely on the amount of label noise. 

We believe the initial transient is likely due to all outputs (including the expected classes) being assigned a probability close 
to 0 by the network at initialization (which would be consistent with the loss being above that of the uniform distribution). However, the reasons for the subsequent increase in coherence are not clear. We discuss in more detail later.

{\bf The Overall Evolution.} If we combine the data from early training (after the initial transient) with the rest of training (e.g., in row 3 we jointly view columns 1 and 2), we find a remarkably consistent pattern across all noise levels: $\mcoherence$ follows a broad parabolic trajectory (albeit with some local variation and noise) where it starts small (around 1), increases to a maximum, and then decreases back to 1.
Thus, there is always an initial increase in coherence, just on different timescales.

{\em From this point of view, the evolution of coherence in the memorization case does not look 
fundamentally different from that in the well-generalizing case.} 

{\bf Impact of Layers.} The bottom 3 rows of Figure~\ref{fig:resnet_full} shows coherence by layer for 3 illustrative layers (the first convolution layer, a convolution layer in the middle, and the final fully connected layer).
Although the specific values are different across the layers,\footnote{Since, 
the interpretability of $\mcoherence$ allows for meaningful comparisons between layers, we 
can get some additional insight into the dynamics of training by studying these values. We do so in the Appendix.}
we notice that the broad trajectory observed for the coherence of the entire model holds for each individual layer. Thus, the trajectory (and in particular the increase) is not driven by one specific layer.

{\bf Impact of Architecture.} We studied one other architecture, Inception, and found a similar increase in coherence and indeed a similar broad trajectory (Figure~\ref{fig:inception}). We note here that Inception does not show the strong initial transient seen in ResNet. 

{\bf Coherence on Test Set.} For completeness, we also measured the coherence on $m$ examples not used for training (drawn from the ImageNet validation set). They are shown in columns 1 and 2 of Figure~\ref{fig:resnet_full} as ``test.'' We defer the discussion to the Appendix.

{\bf Reconciliation with Other Studies.} Finally, we note that it is difficult to compare our experimental results with \citet{Fort19} and \citet{Sankararaman19}, since we use different metrics, sample sizes, datasets (ImageNet v/s {\sc cifar}), and study different effects. But, in as much as they can be compared, we did not find contradictions. Please see the Appendix for more details.

\section{Discussion and Future Work}

% In future work it would be interesting to study other networks and on other datasets to see if the initial increase in coherence happens there as well. It would also be interesting to understand the effects of momentum, augmentation, weight decay, learning rate changes, etc. on gradient alignment through the lens of $\mcoherence$.

% TODO: Other future work: Coherence can be used to freeze layers, etc. Impact of weight decay, dropout, no momentum, Coherence for outlier detection, layers and coherence

{\bf Coherence, Generalization and CG.} 
At a high-level, our experiments provide additional evidence for the connection between the alignment of per-example gradients and generalization~\citep{Fort19, Chatterjee20}. But as our data shows this connection is complicated.

According to CG, the generalizability of SGD is an ``inductive invariant'' that 
the transition dynamics attempts to maintain at each step (as far as possible, given the coherence at that time). 
Thus, if early on in training, there is low coherence (causing the inductive invariant to be violated) then all bets about generalization are off even if there is relatively high coherence later on. 
From this perspective, the high coherence observed in the random label case does not contradict CG. 
 
At the same time, the low coherence seen on real data after the initial transient may be viewed as contradicting the theory. However, we conjecture that since it only persists for a relatively few steps (about 100 steps compared to many epochs for random) there is not enough time for overfitting to occur. Similarly, the extended period of low coherence after the data has been fit does not totally destroy generalization since by that time the gradients are small.

Therefore, it is interesting to consider metrics to predict generalization that are derived from coherence but account for the distance travelled in parameter space. 
Finally, an important test of CG comes from causal interventions to suppress ``weak'' gradient directions (directions supported by a few examples) which are shown to prevent overfitting (e.g. winsorized gradients~\citep{Chatterjee20} and RM3~\citep{Zielinski20}). 
It would be interesting to study these through the lens of $\mcoherence$.% or derived metrics.

%\vfill
%\phantom{M}

{\bf Evolution of Coherence.} 
The evolution of coherence appears to be controlled by two opposing forces. 
On the one hand, as training progresses and examples get fit, coherence is consumed (as per Lemma~\ref{lem:zero} and the Coherence Reduction example). 
On the other hand, as our experiments show, coherence is 
also created during training. 

One may imagine an uneasy equilibrium between these opposing tendencies leading to expansion and contraction in coherence. As soon as significant coherence builds up, it leads to an increase in the effective learning rate (higher relative gradient norm) leading to faster consumption. 
Ultimately, of course, consumption wins out since a stable state is only reached when the gradient becomes small, falling to the orthogonal limit (if the system is sufficiently over-parameterized and we are in the interpolation regime) or below (if under-parameterized and improving loss on one example can only come at the cost of another). 

%\vfill
%\phantom{M}

{\bf Separation of Generalization and Optimization.} 
Optimization and generalization are difficult to disentangle in Deep Learning, but our 
observations point to a possible separation of concerns.

CG provides a uniform {\em first-order} explanation of memorization and generalization based on the simple observation that each step of SGD preferentially reduces the loss on multiple examples if such directions exist, i.e., coherence (locally at a step in training) leads to generalization (locally at that step). But CG does not explain where the coherence comes from, other than to say it depends on the dataset and the model.

Our experiments show that SGD on neural networks, not just exploits coherence, but creates it. Since this creation happens even with random labels where there is nothing to learn (i.e., no generalization), there is reason to believe that this creation is purely an optimization phenomenon. 
Going back to the analogy with random forests, the creation of coherence is similar to the finding of commonality (possibly spurious) between examples during decision tree construction. 
 
Therefore, we believe, what is required is to augment the first-order understanding of generalization provided by CG with a {\em second-order} theory of optimization that explains how coherence or gradient alignment is created.
To that end, the dramatic difference in growth rate of coherence between real and random labels suggests a compounding effect that amplifies existing coherence (perhaps similar in spirit to Lemma~\ref{lem:scale} and mini-batch amplification).
Understanding this process is an important area of future work for us.

\subsubsection*{Acknowledgments}

We thank Sergey Ioffe, Firdaus Janoos, and Alan Mishchenko for many interesting and stimulating discussions on this topic. We thank Michele Covelle, Shankar Krishnan, and Rahul Sukthankar for reviewing early drafts of this paper.

\bibliography{paper.bib}
\bibliographystyle{iclr2021_conference}

\appendix

\section{Appendix: Additional Experimental Results and Discussion}
\label{sec:appendix}

{\bf Experimental Setup.} Our code for running experiments was heavily based on an open source Tensorflow example,\footnote{https://github.com/tensorflow/tpu/tree/master/models/experimental/resnet50\_keras} with modifications to allow label randomization and coherence metric logging. We used SGD with momentum (0.9), a batch size of 4096, and the learning rate schedule proposed in~\cite{Goyal17}. We did not use weight decay or random augmentation of the input. Image size used for Inception-V3\footnote{https://www.tensorflow.org/api\_docs/python/tf/keras/applications/InceptionV3} experiments was $299 \times 299$.

{\bf Variation across Layers.} One advantage of $\mcoherence$ is that it is natural to use it to compare different projections of the per-example gradients and as such can be used to directly compare different layers with each other.
Rows 4, 5 and 6 of Figure\ref{fig:resnet_full} shows the $\mcoherence$ of the classification layer, a convolution layer in the middle and the first convolution layer respectively. We only show the $\mcoherence$ for the weights (since in a spot check the $\mcoherence$ for weights and biases for a given layer looked very similar).
We make a few observations.

First, convolutional layers have higher coherence than the fully connected layer.
Although this could be a function of depth, we note that this may also be expected for a different reason. Since convolutional layers have filters that are instantiated at multiple sites, and the gradients from those sites for a single example add up in the overall gradient for a single gradient. Therefore, by reasoning similar to that of Lemma~\ref{cor:amp}, we expect the coherence of the gradient across sites to be greater than those for the sites individually. And with more sites, we expect greater coherence.
This is another way to see that weight sharing prevents overfitting.

Second, the convolutional layers, particularly the first one shows high coherence for random labels (though still generally lower than those for real labels). However, in the fully connected layer, there is a much greater difference between real and random labels. 
For random, it always is less than 10, whereas for real, it is usually above 10 till accuracy reaches 100\%. And for much of the time till epoch 5 when overfitting starts, it is above 100 or even 1000.

Third, the only place where $\mcoherence$ falls consistently and significantly below 1 (the orthogonal limit) is in the first convolutional layer for real data {\em after} training accuracy has reached 100\%. 
At that point, the layer may be over-constrained, i.e., improving the loss on one example may degrade it on another (though see the discussion below on test set coherence).
Everywhere else, $m$-coherence tends to be at or above the orthogonal limit in line with our expectation that this learning problem is over-parameterized.

\begin{figure}[t]
\centering
\includegraphics[width=1\textwidth]{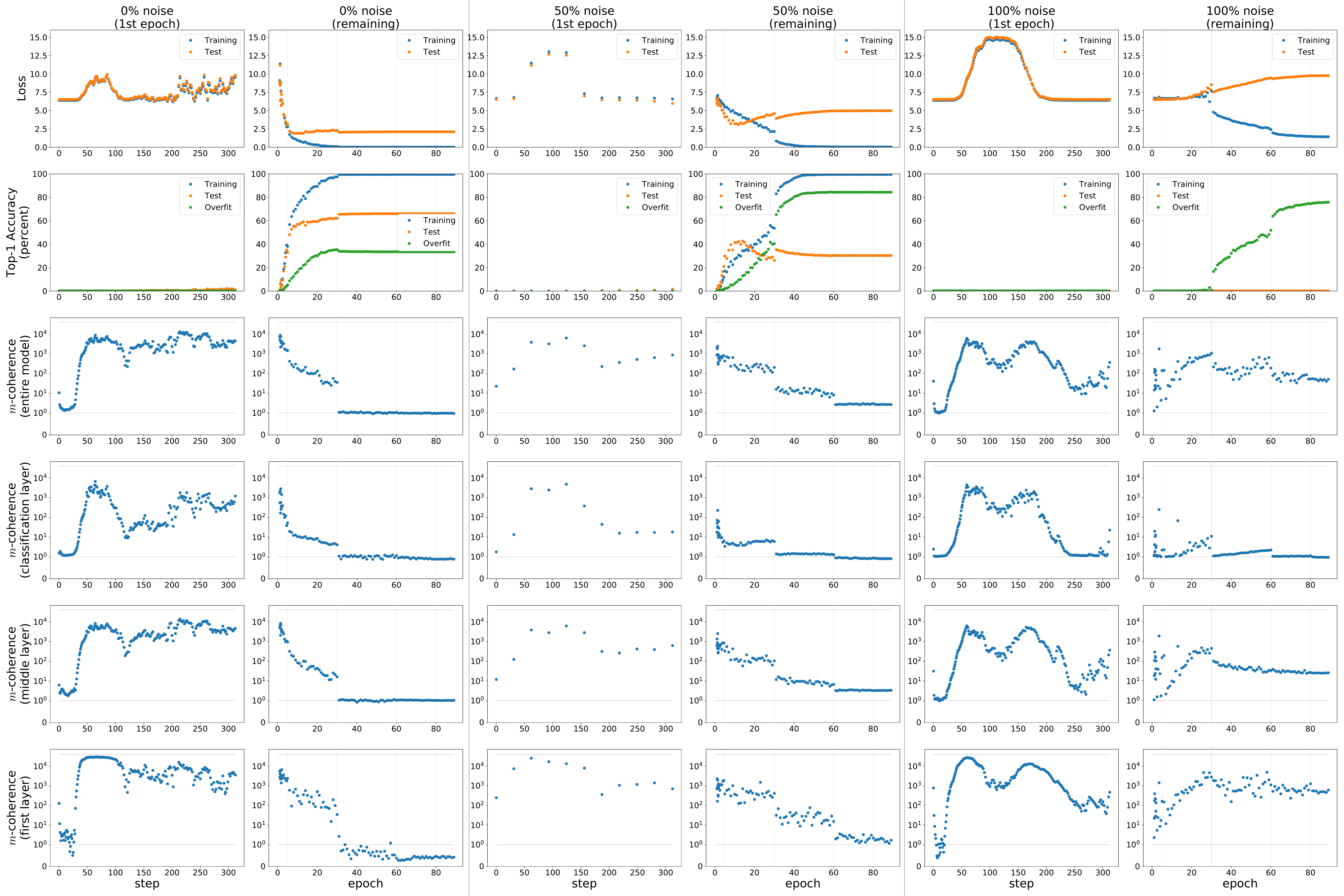}
\caption[foo]{The evolution of alignment of per-example gradients of a Inception-V3 network during training as measured with $\mcoherence$ on samples of size $m=40,356$ on 3 variants of ImageNet with different amounts of label noise. 
We note that the results are qualitatively in agreement with what we see on 
ResNet, i.e., in both real and random cases we see coherence increase.\footnotemark~Horizontal lines for $m$-coherence are shown at 1 (the orthogonal limit) and at $m$. Vertical lines indicate sharp reductions in learning rate. 
For the 1st epoch of 50\% noise, in order to save compute, we only computed $m$-coherence at initialization and at 10 other points during the epoch.}
\label{fig:inception_full}
\end{figure}

Finally, we note that the different layers for Inception in Figure~\ref{fig:inception_full} show similar characteristics as those for ResNet.

\footnotetext{One interesting difference is that the initial transient seen in ResNet happens a little later with Inception but leads to similar high coherence across all 3 variants (though it is most pronounced with random labels). 
We are not sure what causes this, but we believe this is a similar transient as ResNet since the loss increases for a short time going beyond the ${\rm ln}(1000) \approx 6.9$ level expected from an uniform distribution at the outputs. One reason for this could be a higher early learning rate
than is appropriate---we used the same learning rate schedule for Inception as we did for ResNet.
}

{\bf Coherence on Test Set.} 
For completeness, we also measured the coherence on $m$ examples not used for training (drawn from the ImageNet validation set). They are shown in columns 1 and 2 of Figure~\ref{fig:resnet_full} as ``test.''
In the first epoch, we find that test and training coherence are roughly similar. However, when we look at the rest of training, we find that in the early part of the rest, test coherence is below that of training coherence, but in the later part, the opposite holds.
This may be further evidence that coherence creation is a pure optimization phenomenon (as per the discussion in Section 6 of the main paper), i.e., the coherence creation (and subsequent consumption) is specific to the training examples.

It is interesting to observe that particularly for the convolutional layers, at the end of training, the test $\mcoherence$ is at 10 whereas training $\mcoherence$ is at 1 or even lower. This may suggest that those layers are indeed very adapted to the idiosyncrasies of the training data to the extent that no further reduction in loss of any of the training examples is possible locally, though they likely have enough capacity to fit new data (though see discussion above for the third observation on layers).

\begin{figure}[t]
\centering
\includegraphics[width=0.99\textwidth]{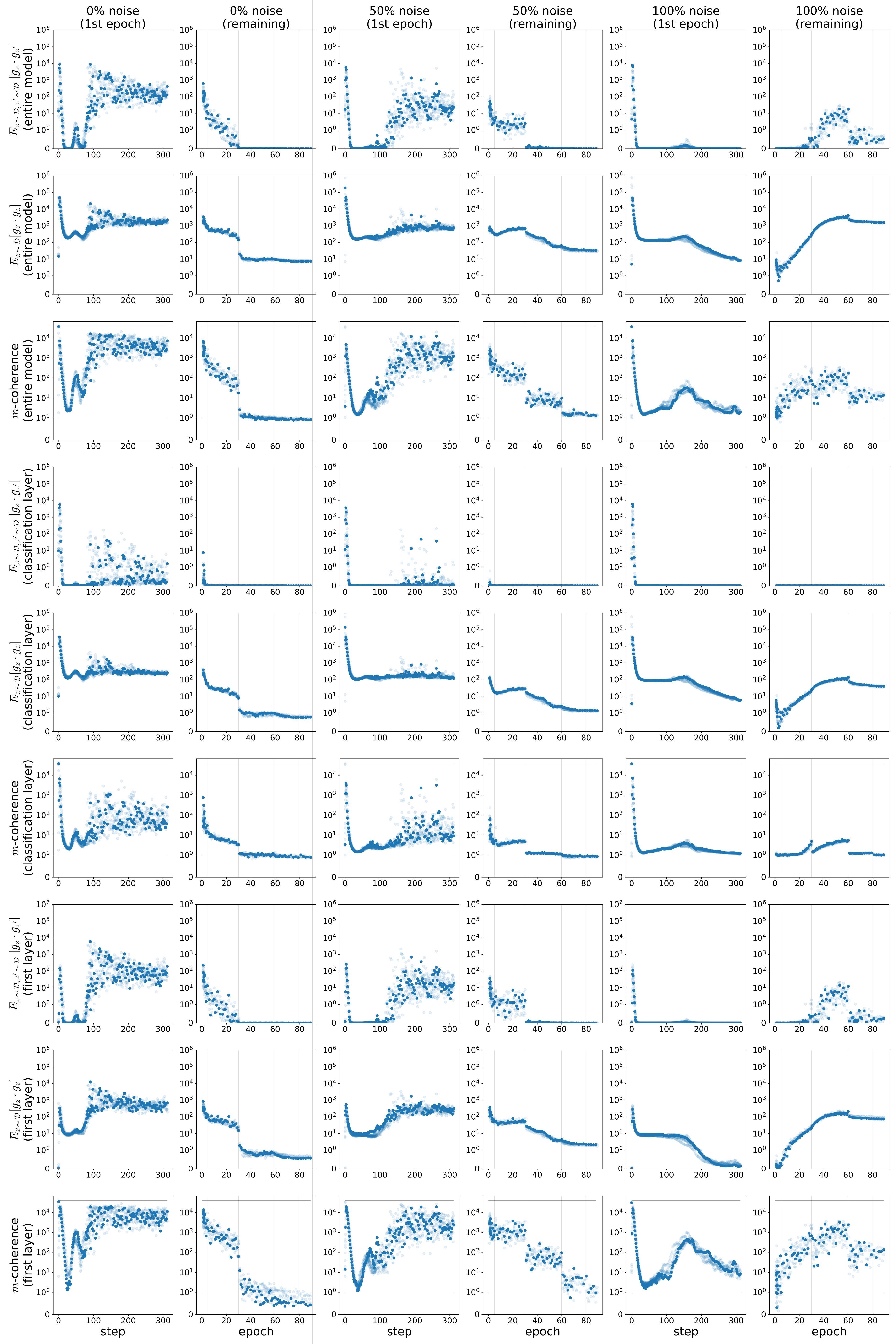}
\caption{The expected gradients in the numerator and denominator for $\alpha$ ({\em not} $m$-coherence) corresponding to Figure~\ref{fig:resnet_full}. 
Note that even when the expected gradient is flat (as may be inferred even from the slope of the loss function), there is activity in the denominator which gets picked up with $\alpha$ or $m$-coherence particularly, if the scale is set appropriately w.r.t. to the orthogonal limit. 
}
\label{fig:gradients}
\end{figure}

{\bf Reconciliation with Other Studies.} 
%
% It is difficult to compare our experimental results with \citet{Fort19} and \citet{Sankararaman19}, since we use different metrics, sample sizes, datasets, and study different effects. 
%
\citet{Fort19} use the cosine and sign stiffness measures to study how gradient alignment depends on class membership, distance in input space between data points, training iteration and learning rate.
They use {\sc mnist}, Fashion {\sc mnist}, {\sc cifar-10/100} and {\sc mnli} datasets. Typical sample sizes are around 500 (for the 10 label datasets) and 3000 (for the 100 label datasets). They do not study label noise or memorization explicitly.
In their class-based analysis, they find that initially, an example of a class only helps other examples in its class and adversely impacts examples of other classes. However, in the course of training, this effect goes down, and stiffness between classes goes up (though only to end up at 0).

We do not explicitly perform a class-based analysis, since with 1000 classes and about 1.2M training examples in ImageNet, we expect on average only 2 to 3 examples in each class pair.
However, implicitly, our study is an inter-class analysis (though {\em not} a class-pair analysis) since in our sample, each example is expected to see roughly 1000 times as many examples of other classes as it does its own class.
Our results indicate that examples in one class do help examples in other classes at different points in training since $m$-coherence is often in 1000s, and in a sample of approximately $40k$, we expect only about 40 examples per class. 

However, since our metric is very different (as discussed in detail in Sections 2 and 3 of the paper), and the error bars in their study are large (as indicated in their Figure 5), we do not directly see any contradictions in the experimental data between their study and ours.
%
%TODO: Discuss their conclusion on generalization in Section 5 ?
% We are probably in a contradiction since coherence can be high even when there is no generalization as we see in the random label case, and also in the real label case.
%
%But probably not worth talking about. 
%
Finally, we do not study coherence as a function of input distance between examples or of learning rate changes, though we are interested in investigating the latter in future work.

\citet{Sankararaman19} show theoretically that high gradient confusion impedes the convergence of SGD, and also analyze how factors such as network depth and width and initialization impact gradient confusion.
They validate their theoretical results with experiments on {\sc mnist}, {\sc cifar-10}, and {\sc cifar-100} (real labels only, since they do not study memorization) where they measure the minimum cosine similarity (MCS) between different training examples (though as discussed in the main paper they compute this over mini-batches of size 128 rather than on individual examples).
They mainly focus on the MCS value at the end of training as various architectural parameters are varied, but in Figures 7(c) and 8(c) in the appendix, they show the trajectory during training.
There, we find that MCS starts low, increases to a peak and then comes back down again, in qualitative agreement with our findings (though we reiterate that our metric is very different).

Finally, we show the individual terms for the numerator and denominator for $\alpha$ in Figure~\ref{fig:gradients}. Although the numerator can be estimated from the slope of the loss curve (as per equation (1) in the main paper), without the denominator to give it scale, it is hard to understand what variations are meaningful. 
As an extreme example, we see that, as expected from the loss curve (and this may be seen in the loss plots from other studies such as \citet{Zhang17, Zielinski20}), in the 100\% random case, for the first 20 epochs or so, the norm of the expected gradient is close to zero (Figure~\ref{fig:gradients}, last column, row 1). 
However, there is significant activity in the denominator (row 2; and this is not typically recorded in experiments). By considering the quotient, and furthermore, {\em by putting it into context with the orthogonal limit} (as we do with $m$-coherence where that limit sets the scale), we can see that there is a definite build up in coherence in that period (row 3).

% TODO: Talk about how our plots are more interesting, -1, 0, 1 mapped to 0, 1/m m and the latter is where most of the action happens.
%

\begin{figure}[t]
\centering
\includegraphics[width=0.45\textwidth]{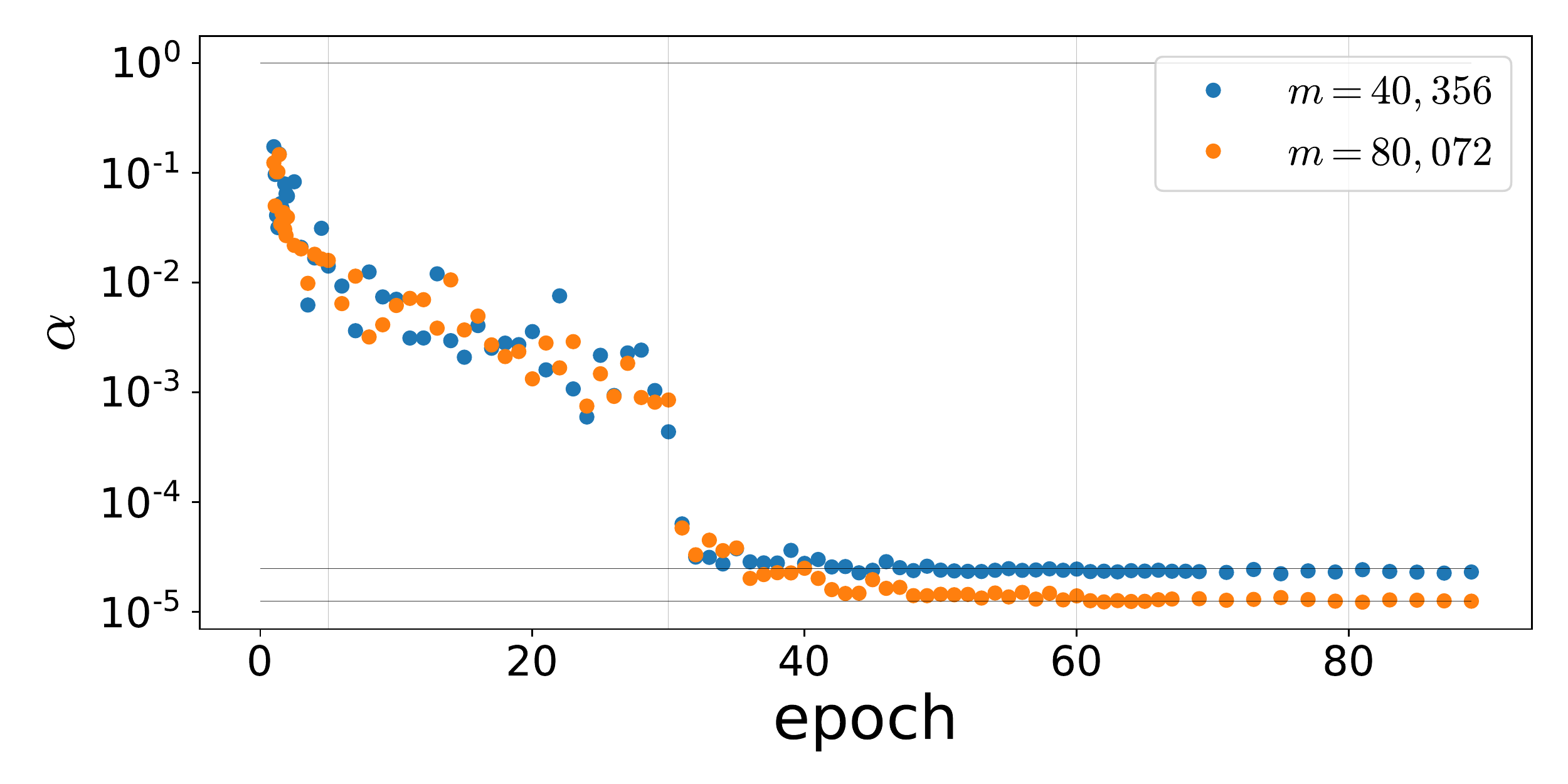}
\hspace{1cm}
\includegraphics[width=0.45\textwidth]{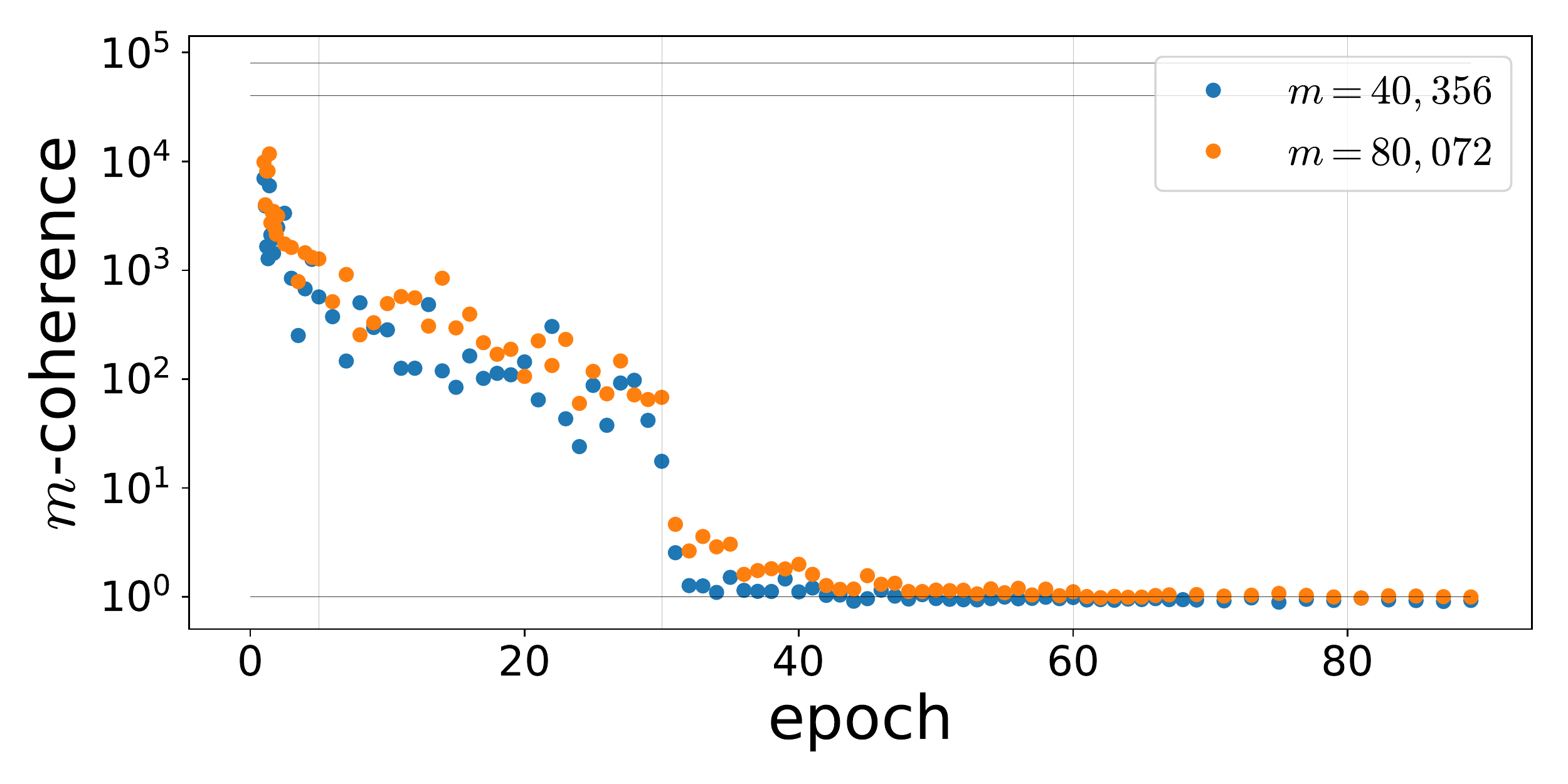}
\caption{To understand the effect of $m$ for the values of $\alpha$ and $\mcoherence$, we plot these values for $m=40,356$ and $m=80,072$ for the ResNet-18 training on 0\% noise.
In both plots we show horizontal lines for the orthogonal limit (which is different for the two samples in the $\alpha$ plot since it is $1/m$, but the same in the $m$-coherence plot since it is 1 in both cases) and the perfect alignment case (which is the same in the $\alpha$ plot since it is 1, and is different in the $\mcoherence$ plot since it is $m$.)}
\label{fig:40vs80}
\end{figure}

{\bf Effect of sample size $m$.} Figure~\ref{fig:40vs80} shows the effect on $m$-coherence and $\alpha$  of approximately doubling $m$ from our baseline value of $40,356$.
We see that the numerical values are generally the same, showing a slightly upward bias in $\mcoherence$ for the larger value of $m$ as might be expected.

\begin{figure}[t]
\centering
\includegraphics[width=\textwidth]{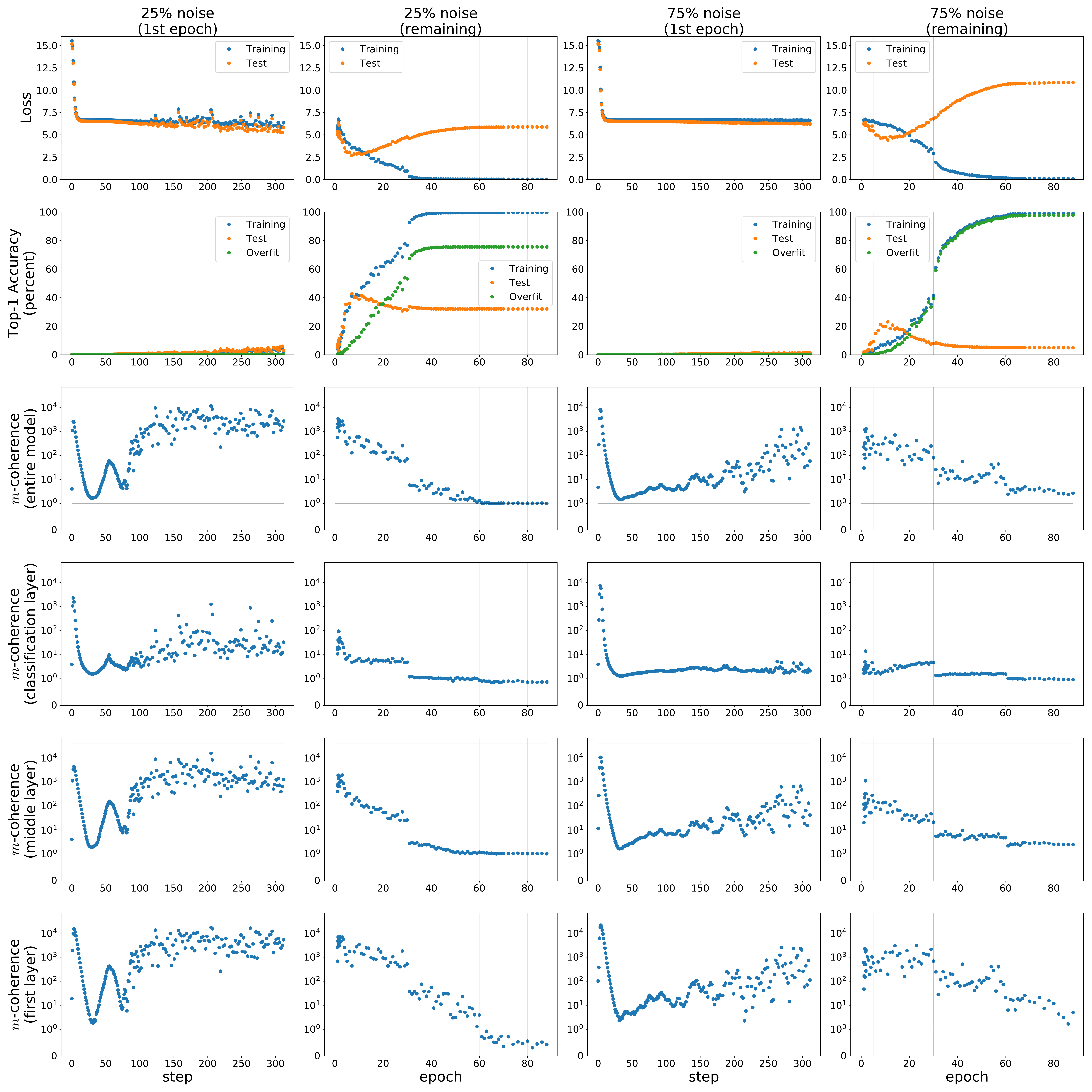}
\caption{
Evolution of $\mcoherence$ for 25\% and 75\% label noise (under the same settings as Figure~\ref{fig:resnet_full}).
This confirms the pattern discussed in the main text that with increasing noise, the rate at which coherence is created in early training slows down.}
\label{fig:resnet_2575}
\end{figure}

{\bf 25\% and 75\% label noise.} 
The data for ResNet-18 training on 25\% and 75\% label noise is shown in Figure~\ref{fig:resnet_2575}. This confirms the pattern noted in the main paper (Section 5) that with increasing noise, the slope with which coherence increases depends inversely on the amount of label noise.

\end{document}